\documentclass[a4paper,twoside]{article}

\usepackage{epsfig}
\usepackage{subfigure}
\usepackage{calc}
\usepackage{amssymb}
\usepackage{amstext}
\usepackage{amsmath}
\usepackage{amsthm}
\usepackage{multicol}
\usepackage{pslatex}

\usepackage{csquotes}
\usepackage{dirtytalk}
\usepackage[hyphens]{url}
\usepackage{natbib}
\let\bibhang\relax
\usepackage{apalike}

\usepackage{enumerate}  
\usepackage{balance}

\usepackage{fancyvrb,xcolor}

\usepackage{color} 
\definecolor{mygreen}{RGB}{28,172,0} 
\definecolor{mylilas}{RGB}{170,55,241}

\usepackage{SCITEPRESS}   

\graphicspath{ {Figures/} }

\begin{document}

\title{Anomaly Detection for Industrial Big Data}

\author{
\authorname{Neil Caithness and David Wallom}
\affiliation{
Oxford e-Research Centre, 
Dept. of Engineering Science, 
University of Oxford, 
Oxford, United Kingdom}
\email{\{neil.caithness, david.wallom\}@oerc.ox.ac.uk}}


\abstract{As the Industrial Internet of Things (IIoT\footnote{\url{http://www3.weforum.org/docs/WEFUSA_IndustrialInternet_Report2015.pdf}}) grows, systems are increasingly being monitored by arrays of sensors returning time-series data at ever-increasing `volume, velocity and variety'\footnote{\url{https://aip.scitation.org/doi/10.1063/1.4907823}} (i.e. Industrial Big Data\footnote{\url{http://www.vix.com.pl/wp-content/uploads/the_rise_of_industrial_big_data.pdf}}). An obvious use for these data is real-time systems condition monitoring and prognostic time to failure analysis (remaining useful life, RUL). (e.g. See white papers by Senseye.io, \textit{Prognostics - The Future of Condition Monitoring}\footnote{\url{https://www.senseye.io/white-papers-senseye-scalable-predictive-maintenance/}}, and output of the NASA Prognostics Center of Excellence (PCoE\footnote{\url{https://ti.arc.nasa.gov/tech/dash/groups/pcoe/}}).) However, as noted by Agrawal and Choudhary\footnote{\url{https://aip.scitation.org/doi/full/10.1063/1.4946894}} `Our ability to collect ``big data'' has greatly surpassed our capability to analyze it, underscoring the emergence of the fourth paradigm of science, which is data-driven discovery.' In order to fully utilize the potential of Industrial Big Data we need data-driven techniques that operate at scales that process models cannot. Here we present a prototype technique for data-driven anomaly detection to operate at industrial scale. The method generalizes to application with almost any multivariate dataset based on independent ordinations of repeated (bootstrapped) partitions of the dataset and inspection of the joint distribution of ordinal distances.}

\onecolumn \maketitle \normalsize \vfill

\section{\uppercase{Introduction}}
\label{sec:introduction}

\noindent Data from energy smart meters represents an appropriate use-case for demonstrating a new method for detecting anomalies in Industrial Big Data, specifically where accurate system process models do not exist and where data-driven techniques are sought instead. The electricity network is a system being monitored by arrays of sensors (smart meters), and represents appropriately high volume, velocity, and variety of data \citep{de_Mauro___2015}. 

Energy theft by meter tampering represents a significant threat to the UK energy industry. The creation of new opportunities for cyber theft and fraud introduced by the imminent deployment of up to 50M smart meters across the UK industry has prompted a search for a data-driven solution to detecting possible theft or fraud events. In the DIET project (Data Insight against Energy Theft\footnote{Innovate UK, Grant Reference Number 102510.}) we developed a new technique using unsupervised machine learning for the identification of anomalous events based on data collected exclusively from smart meters.

Electricity smart meters typically collect standardized consumption data (kWh in 48 half-hour bins per day) and an associated stream of non-standardized or meter specific \say{event} or \say{logging} data. Event streams may consist of nominal codes that track the actions on the smart meter and may relate to activities such as accessing the cache via modem, turning the power supply off, or many more of increasingly technical nature. It is yet unknown whether meter tampering for theft, fraud, or otherwise, carries with it specific and detectable event sequence signatures (i.e. there is no relevant process model in existence.)

The question then becomes, can theft or meter tampering be credibly detected from the data streams alone, without reference to defining event sequence signatures? Without credible and/or sizable training sets of theft or fraud cases with which to train supervised machine learning systems, we position this task instead in the context of anomaly detection, using unsupervised machine learning techniques. 

This is closely related to the field of outlier detection in the data science literature, where, in a well-known definition by \cite{Hawkins_1980} an outlier is 
\begin{displayquote}
\it an observation which deviates so much from other observations as to arouse suspicion that it was generated by a different mechanism.
\end{displayquote}

\noindent Or,  according to \cite{Barnett__1994}
\begin{displayquote}
\it an observation [...] which appears to be inconsistent with the remainder of that set of data.
\end{displayquote}

A resurgence of research activity on outlier detection algorithms was triggered by the seminal work of \cite{Knorr__1997}, and new models have continued to be developed \citep{Aggarwal_2013,Akoglu___2015,Chandola___2009,Schubert___2014}, though none has achieved universal acclaim as being applicable and effective in all cases.


\cite{Campos___2016} have systematically and thoroughly reviewed the literature and development of unsupervised outlier detection algorithms and found that \say{little is known regarding the strengths and weaknesses of different standard outlier detection models}, that \say{the scarcity of appropriate benchmark datasets with ground truth annotation is a significant impediment to the evaluation of outlier methods}, and still further that \say{even when labeled datasets are available, their suitability for the outlier detection task is typically unknown}. Thus endeavors to develop and critically evaluate new unsupervised techniques are both timely and warranted.

We refer to the general concept of \textit{outlier} instead as \textit{anomaly}, to emphasize the suspected origin in a different underlying mechanism as characterized by \cite{Hawkins_1980} cited above, and to distinguish it from outliers in the normal tail of a statistical distribution produced by a single unified mechanism. In this paper we present a new approach to the task of anomaly detection specifically designed to accommodate datasets of different data-types on the same set of cases (e.g. consumption and event data for a portfolio of smart meters). This variant of the method is based on independent analyses of two data streams to establish case-by-case measures of outlierness (distance). Identification of potential anomalous cases is then achieved by inspecting the joint-distribution of distances derived from a novel density analysis. This joint-distribution is susceptible to new and special statistical interpretation that we introduce here, but which we will develop further in subsequent publications. Our contention is that in this application, meter tampering for theft, fraud, or otherwise, will insert into the dataset cases that are not representative of the background mechanism of the system as a whole.

Finally, we suggest that this procedure generalizes to anomaly detection for almost any multivariate dataset with any combination of data-types by applying independent analyses to two or more (possibly randomized) partitions of the dataset and then inspecting the joint distribution of cases.

\section{\uppercase{Method}}
\label{sec:method}

\begin{figure}[!htbp]
  \centering
  \includegraphics[width=1.0\linewidth]{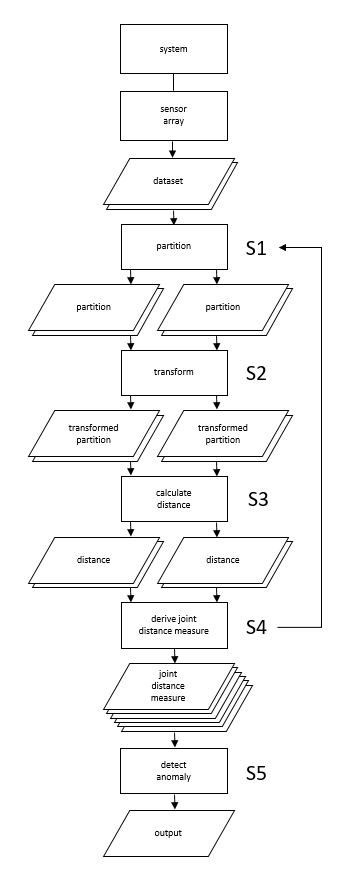}
  \caption{Schematic diagram of the method workflow.}
  \label{Fig:Schematic}
\end{figure}

\noindent For any system (here the electricity network) being monitored by technical equipment (here a sensor array comprising the set of smart meters) that generates a multivariate cases-by-variables dataset, we devise the following method of analysis. It is useful here to refer to a schematic diagram of the workflow of the method as shown in Figure \ref{Fig:Schematic}. We describe in detail the five processes labeled S1-S5.

\subsection{Partition (S1)}
\label{subsec:partition}

The first novel aspect of the method is that datasets to be studied are always split into two or more partitions, vertically by variables, and each partition is analyzed separately. Cases need not be present in all partitions (or a case may have null data across the variables in a partition) but anomalous cases can be detected only from the set of cases in common across all partitions.

There are a number of criteria for partitioning:

\subsubsection{Different data types}
\label{subsec:different_data_types}

The smart meter event/consumption dataset presents time series data that is a mixture of data-types: individually time-stamped nominal-scale event codes; and time-binned ratio-scale consumption data. Where the downstream analysis requires different variants depending in the data-types of the variables, then partitioning on data-type is a natural evolution of the workflow.

\subsubsection{Random partitioning}
\label{subsec:random_partitioning}

Where partitions do not require different kinds of analysis we make partitions at random and repeat many times (a process with similar benefits to statistical bootstrapping \citep{Efron__1994}).

\subsection{Transform (S2)}

There are two steps in the transform process: coding and ordination.

\subsubsection{Coding}

For the nominal-scale event data we first transform by frequency count over some specified time period. We found by inspection that three months is an effective period: shorter and many events have zero count, so this is effectively ignoring potentially relevant data; longer and the anomalous data produced by the event of a meter tamper could be so diluted by the background of normal operation as to go undetected. An alternative treatment for nominal-scale data would be to apply a numeric coding. If the partitions are made at random and repeated sampling is performed, and if randomized numeric coding is applied with each repetition, then the association between codes remains unbiased.

For the ratio-scale consumption data, we wish to capture a relevant cyclical frequency. There are clear daily, weekly, and seasonal energy consumption patterns, and although these cycles differ between domestic and commercial properties, they remain the dominant patterns for both types of property. In our analysis we recode the standard 48 half-hourly bins per day of consumption data collected by typical smart meters to a total daily consumption, averaged by day of the week. There are many other coding possibilities, and many of them may prove effective in future experiments.

\subsubsection{Ordination}

We use several variants of methods from the family of dimension reduction techniques that employ singular value decomposition (SVD). Specifically, principal components analysis (PCA) and correspondence analysis (CA), depending on the data types of the partitions.

The biplot provides a convenient visualization of the ordination but is not itself integral to the technique. We illustrate our results with various kinds of plots, including biplots, in the following section.

The {\it biplot} \citep{Greenacre_2010} is a graphical device that shows simultaneously the rows and columns of a data matrix as points and/or vectors in a low-dimensional Euclidean space, usually just two or three dimensions. \cite{Greenacre_2013} introduced the {\it contribution biplot} in which the right singular vectors (column contribution coordinates) of a dimension reduction analysis show, by their length, the relative contribution to the low-dimension solution. Contribution biplots can be used with any of the methods that perform dimension reduction by singular value decomposition (SVD), these include correspondence analysis (CA), principal component analysis (PCA), log-ratio analysis (LRA), and various derived methods of discriminant analysis.



SVD is a factorization of a target matrix T such that 

\begin{equation}\label{eq1}
\rm T = U\Gamma V^T
\end{equation}

What distinguishes the various methods is the form of the normalization applied to T before performing the SVD. In CA this normalization is the matrix of standardized residuals

\begin{equation}\label{eq2}
\rm T = D_r^{-1/2}(P-rc^T)D_c^{-1/2}
\end{equation}

\noindent where P is the co-called correspondence matrix P~=~N/n, with N being the original data matrix and n its grand total, row and column marginal totals of P are r and c respectively, and D\textsubscript{r} and D\textsubscript{c} are the diagonal matrices of these.

In the analysis of a cases-by-variables data matrix, the right singular vectors of the SVD, V, are the contribution coordinates of the columns (variables). A further transformation involving a scaling factor D\textsubscript{q}, such that 

\begin{equation}\label{eq3}
\rm F = D_q^{-1/2}U\Gamma
\end{equation}

\noindent defines the principal coordinates of the rows (cases). The joint display of the two sets of points in F and V can often be achieved on a common scale, thereby avoiding the need for arbitrary independent scaling to make the biplot legible. 

The appropriate normalizations and the derivation of scaling factors for the alternative methods are detailed in his Table 2 and in various equations given in \cite{Greenacre_2013}. We use CA for the ordination of the event data, following a double log transform of the frequency data, N\textsubscript{0} such that 

\begin{equation}\label{eq4}
\rm N = ln(ln(N_0+1)+1)+1
\end{equation}

Note that the successive additions ($+$1) in Equation (\ref{eq4}) above are simply to avoid taking ln(0). This is a convenience, introducing an appropriate scaling so as to make the biplot legible, but does not otherwise alter the analysis. For the ratio-scale consumption data we use the PCA method of \cite{Greenacre_2013}, after centering and standardizing the input data by variable.

\subsection{Calculate distance (S3)}
Both ordination techniques, whether CA or PCA, result in a matrix F of principal coordinates of the rows (cases) as in Equation (\ref{eq3}). This matrix has the same number of dimensions (columns) as variables in the raw input data, however the information content of the data is now concentrated towards the higher order components (i.e. towards the left-most columns of F). This is the central purpose of the dimension reduction performed by  SVD, and typically, a scree plot is used to inspect the degree of dimension reduction, essentially a plot of the eigenvalues, $\Gamma$ in Equation (\ref{eq1}). 

A decision needs to be made as to how many components to retain, referred to as a stopping rule \citep{Jackson_1993,Peres-Neto___2005}. A conventional rule is to retain only those components with corresponding eigenvalues $>$1 (known as the Kaiser-Guttman criterion, \citep{Nunnally___1994}), which is the rule we apply here, though this is a tunable parameter of the method and a range of values should generally be explored. Once a stopping rule has been decided the case-by-case distances d from the origin in Euclidean space are calculated for the k number of retained dimensions. This is done separately for each partition. The following code is provided for clarity.

\begin{quote}
{\it e.g. Matlab Code}
\begin{small}
\begin{verbatim}
d = sum(F(:,1:k).^2,2).^(1/2)
\end{verbatim}
\end{small}
\end{quote}

\subsection{Derive joint distance measure (S4)}

The second novel aspect of the method, after partitioning the dataset, is to examine the joint-distribution of distances derived from the separate ordinations of the partitions. Where the partitions have vastly different numbers and types of variables, and where the specific ordination techniques differ between partitions, (as in our case of event data, circa 250 variables of event frequency counts, analyzed by CA, vs. consumption data, seven variables of ratio-scale data, analyzed by PCA), then comparison should be made on the rank order of distances, rather than directly on the distances themselves.

If all the data across all the variables were generated by independent random processes, then there would be no relationship between the rank-ordering of cases in the two lists. If the variables are at least partially correlated (as is usually the case for real-world data) then we would expect a correlation between the rankings derived from the two partitions, but we would still expect an even spread of associations. A scatter plot of the two rank ordered lists will reveal the nature of the association. A correlation among variables will manifest as a concentration of points towards the diagonal, but from a unified underlying process we would not expect much departure from an even spread along the diagonal. If a second, distinct process inserts cases into the dataset, we could expect that these may be manifested as a departure from the uniform density of points, possibly forming locally high density clusters. We plan to develop the statistics of this phenomenon further in subsequent publications.

\subsection{Detect anomaly (S5)}

Following on from the derivation of a joint distance measure based on the density of points in the joint distribution of rank orders, as described in S4, we now standardize the measure and inspect the departure from the mean density in units of standard deviation. Cases at the far extremes of departure from the mean may well be interpreted as being so divorced from the background process generating the bulk of the data as to be anomalies produced by a different mechanism \citep{Hawkins_1980}.

This final process in the method proceeds to find those cases at the far extremes of departure from the mean density, and to report them as likely anomalies that require an alternative explanation.

\section{\uppercase{Data}}
\label{sec:data}

\noindent We analyzed a dataset obtained from British Gas consisting of anonymized records for circa 120,000 smart meters with daily  electricity consumption (kWh in 48 half hour bins) and individually time-stamped event codes as logged by each meter over a three-month period. According to the data confidentiality agreement we are unable to make these data available in supplemental material for this publication and the dataset has been returned to ownership of British Gas.

\section{\uppercase{Results}}
\label{sec:results}

\begin{figure}[!htbp]
  \centering
  \includegraphics[width=1.0\linewidth]{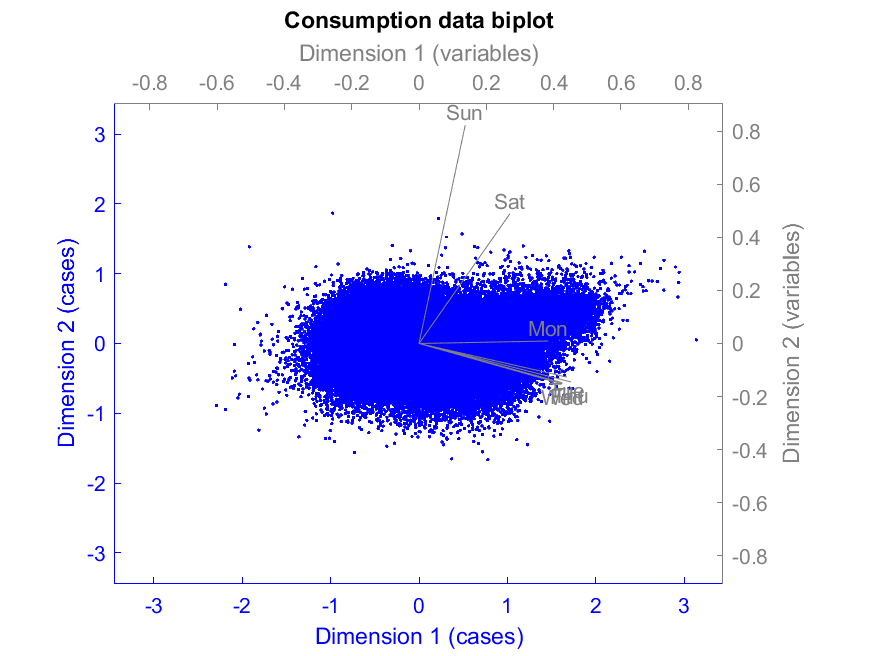}
  \caption{PCA biplot of consumption data.}
  \label{Fig:ConsBiplot}
\end{figure}

\noindent We present results of the analysis in a series of plots. Figure \ref{Fig:ConsBiplot} shows a biplot of the consumption data in the first two dimensions of the PCA ordination. Cases (the blue dots) can be ranked by their distance from the origin in any chosen number of dimensions (up to seven). Interpreting the vectors of the variables (days of the week) we see the weekend being orthogonal to week days as we might expect for small business properties.

The cloud of cases is roughly elliptical in the first two dimensions with clearly identifiable outliers, but none that would necessarily arouse suspicion.

\begin{figure}[!htbp]
  \centering
  \includegraphics[width=1.0\linewidth]{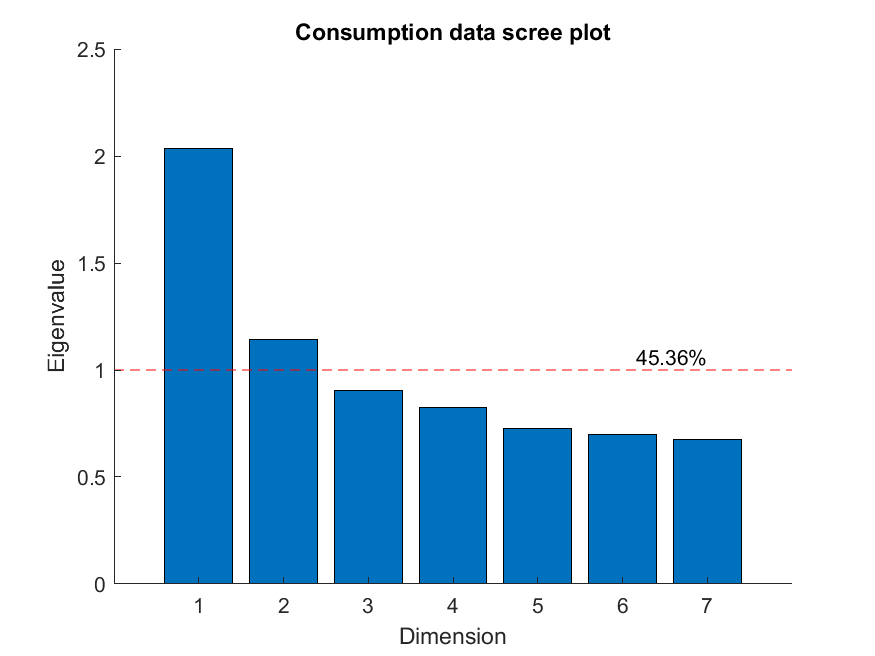}
  \caption{Screeplot of consumption data.}
  \label{Fig:ConsScreePlot}
\end{figure}

Figure \ref{Fig:ConsScreePlot} shows a scree plot of the consumption data ordination. We can be confident that there is significant dimension reduction (the concave shape). The Kaiser-Guttman rule selects only the first two components with eigenvalues $>$1, however these two components account for only 45.36\% of the variance of the original data.

\begin{figure}[!htbp]
  \centering
  \includegraphics[width=1.0\linewidth]{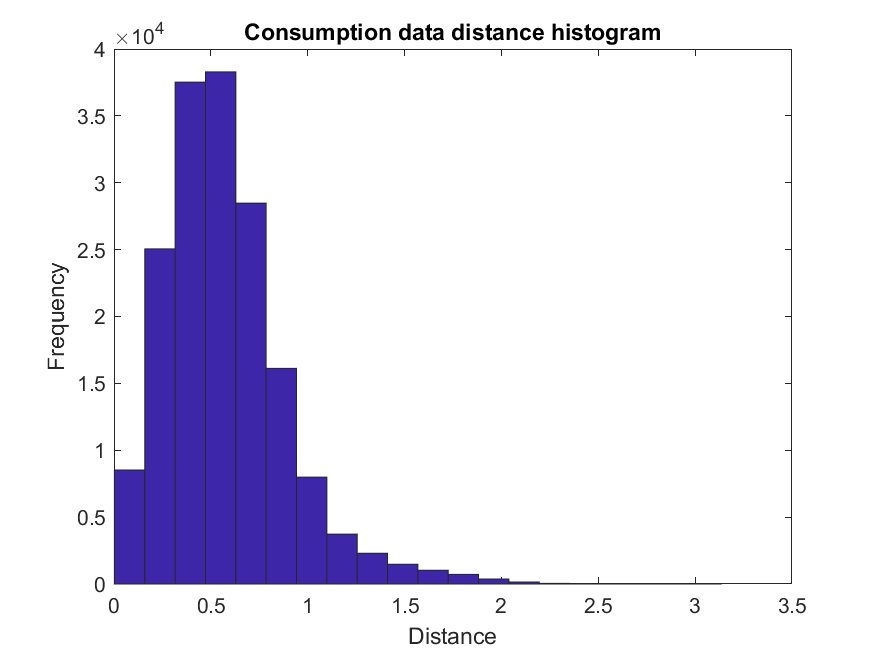}
  \caption{Ordinal distance histogram of consumption data.}
  \label{Fig:ConsDistancePlot}
\end{figure}

Figure \ref{Fig:ConsDistancePlot} shows a histogram of the consumption PCA ordinal distances with a right-skew, and a not exceptionally long tail.

\begin{figure}[!htbp]
  \centering
  \includegraphics[width=1.0\linewidth]{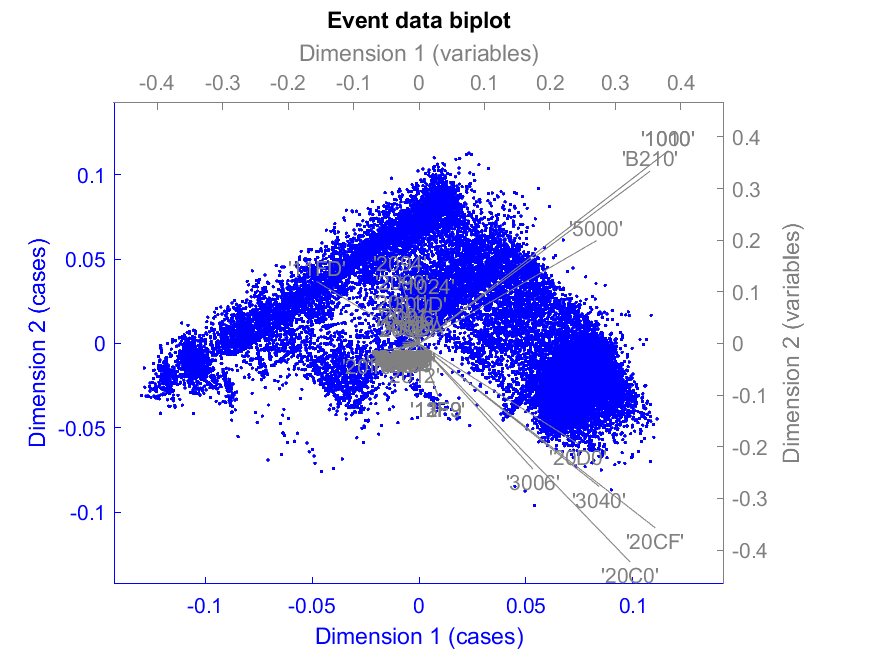}
  \caption{CA biplot of event data.}
  \label{Fig:EventBiplot}
\end{figure}

Figure \ref{Fig:EventBiplot} shows a biplot of the CA ordination of the event data in the first two dimensions. A hand-full of variables dominate the solution, in two orthogonal sets. The rest of the variables contribute only a minor influence on the solution.

\begin{figure}[!htbp]
  \centering
  \includegraphics[width=1.0\linewidth]{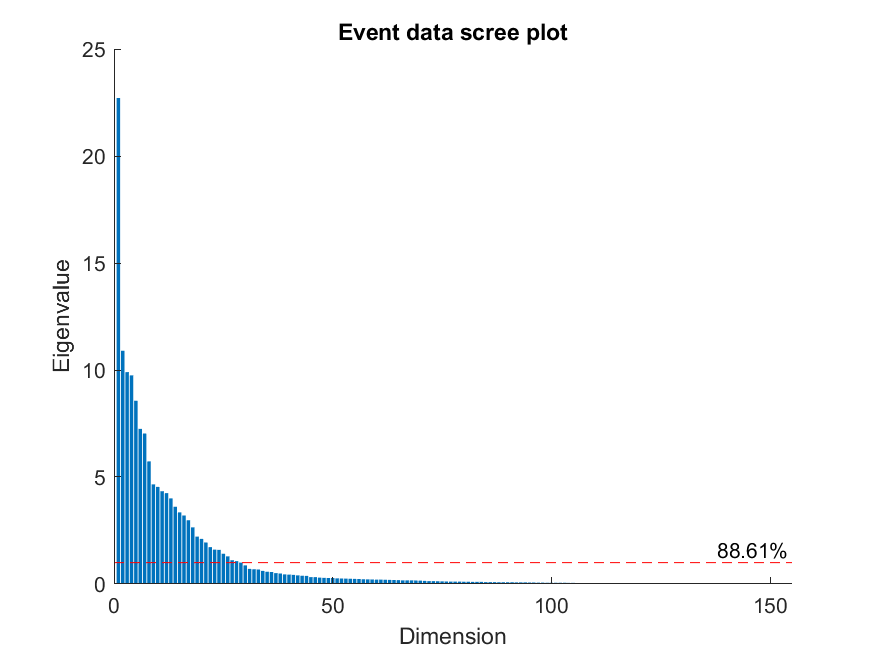}
  \caption{Scree plot of event data.}
  \label{Fig:EventScreePlot}
\end{figure}

Figure \ref{Fig:EventScreePlot} shows a scree plot of the event data ordination. Here we can be even more confident that there is significant dimension reduction. The Kaiser-Guttman rule \citep{Jackson_1993} selects about 20 out of circa 150 variables with eigenvalues $>$1, and these account for 88.61\% of the variance of the original data.

\begin{figure}[!htbp]
  \centering
  \includegraphics[width=1.0\linewidth]{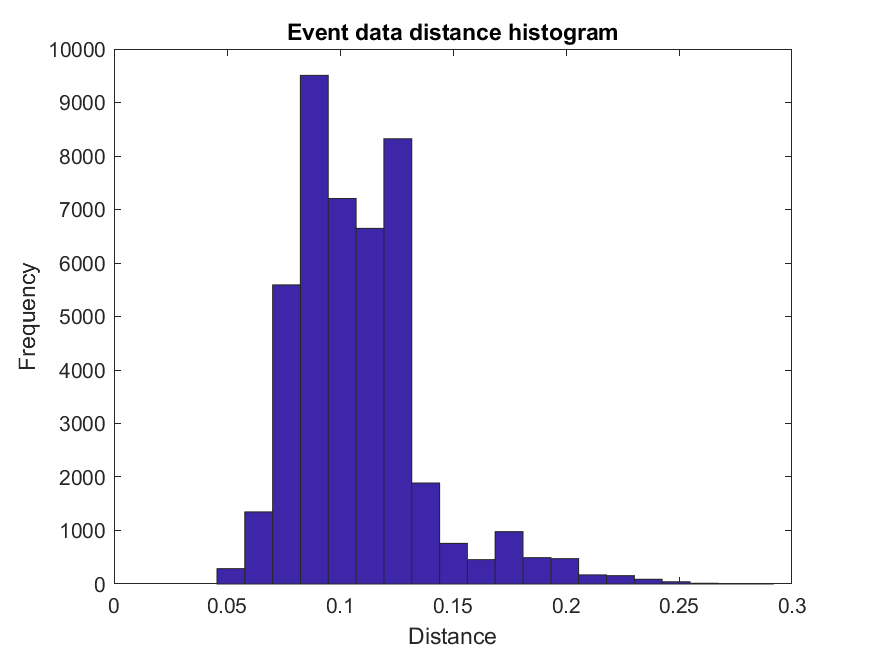}
  \caption{Ordinal distance histogram of event data.}
  \label{Fig:EventDistancePlot}
\end{figure}

Figure \ref{Fig:EventDistancePlot} shows a histogram of the event CA ordinal distances with a right-skew, and again a not exceptionally long tail.

\begin{figure}[!htbp]
  \centering
  \includegraphics[width=1.0\linewidth]{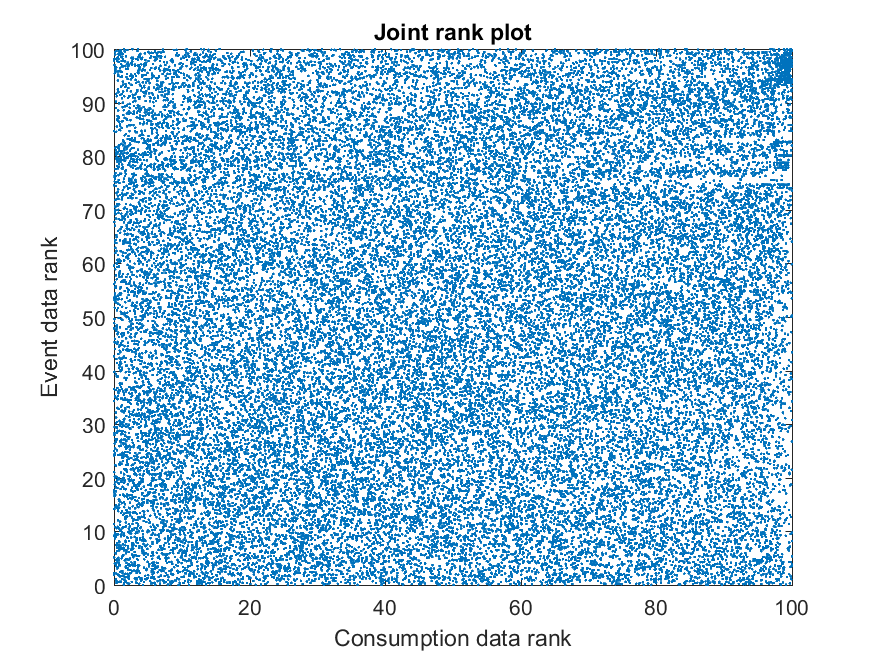}
  \caption{Joint rank scatter plot.}
  \label{Fig:JointRankPlot}
\end{figure}

Figure \ref{Fig:JointRankPlot} shows a scatter plot of the ranked distances from the two ordinations of consumption and event data. This shows, for the most part, that the data in the two partitions are independent (the more-or-less even cloud of data points across the entire space.) If the variables in the two partitions were partially correlated we would expect a concentration in density of points towards the diagonal. 

What we're not expecting from a unified underlying process is (much) variation in density of points along the diagonal. For these data we see a distinct cluster of high density at high ranks in the upper right corner.

\begin{figure}[!htbp]
  \centering
  \includegraphics[width=1.0\linewidth]{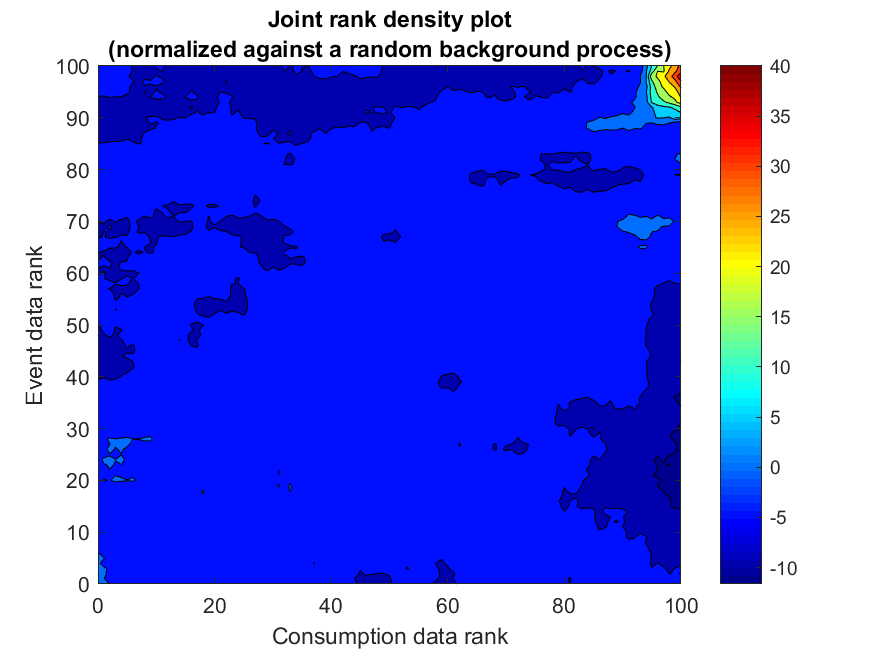}
  \caption{Joint rank density plot.}
  \label{Fig:JointRankDensityPlot}
\end{figure}

Figure \ref{Fig:JointRankDensityPlot} shows the density of the joint ranks as a contour plot where density has been scaled to units of standard deviation. The high density cluster in the upper right corner is from two to 40 standard deviations away from the mean of the background process, and contains no more that 560 cases, out of a total of some 120k cases. A slight concentration towards the diagonal is also evident for the background process. 

\begin{figure}[!htbp]
  \centering
  \includegraphics[width=0.8\linewidth]{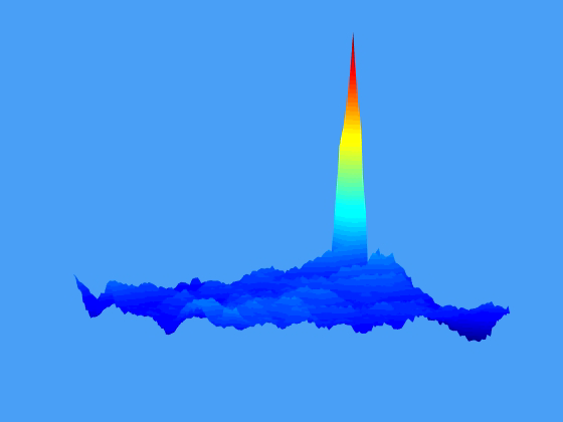}
  \caption{Joint rank density 3-d.}
  \label{Fig:JointRankDensityPlot-3-d}
\end{figure}

Figure \ref{Fig:JointRankDensityPlot-3-d} shows a surface plot using the same scaling in units of standard deviation. We interpret the spike in density to indicate a set of anomalous cases that are probably derived from a different underlying mechanism to the rest of the dataset.

\begin{figure}[!htbp]
  \centering
  \includegraphics[width=1.0\linewidth]{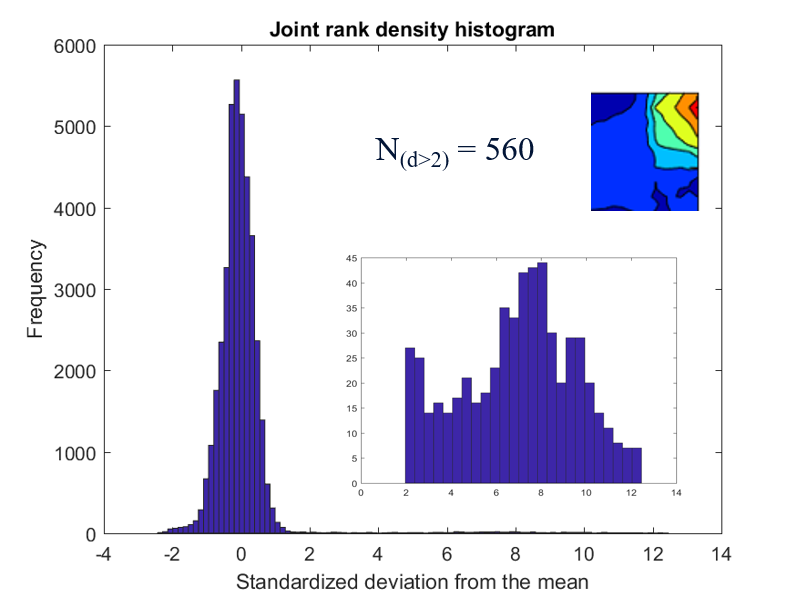}
  \caption{Joint rank density histogram.}
  \label{Fig:JointRankDensityHistogram}
\end{figure}

Figure \ref{Fig:JointRankDensityHistogram} illustrates the long tail of the distribution and shows just 560 cases with std $>$2, out of a dataset of some 120k cases.

\section{\uppercase{Performance}}
\label{sec:performance}

\noindent Computing the SVD in each ordination is the most time consuming element of the method and the most critical obstacle to scaling for Industrial Big Data. In a series of experiments we tested three different implementations and measured the performance as execution time iterating over an increasing number of dataset rows. Matlab code for the three implementations is shown in Sections \ref{subsec:full} to \ref{subsec:vectprized}. Tests were run using Matlab(R) Version 9.3.0.713579 (R2017b) on an ordinary desktop computer with an Intel(R) Core(TM) i7-4790 CPU @ 3.60GHz, 3601 Mhz, and Installed Physical Memory (RAM) of 16.0 GB. 

\begin{figure}[!htbp]
  \centering
  \includegraphics[width=7.5cm]{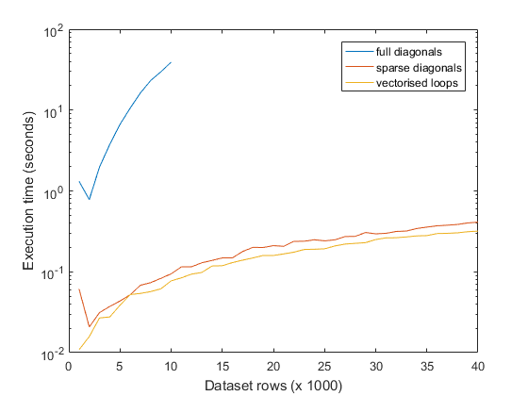}
  \caption{Performance evaluation plot.}
  \label{Fig:PerformancePlot}
\end{figure}

Results of the performance tests are shown in Figure \ref{Fig:PerformancePlot}. The `naive' but straightforward implementation following the mathematical notation and using full diagonal matrices performs poorly, and is unusable for Big Data unless modified. Both the sparse and vectorized implementations perform adequately (vectorized being slightly better). These tests demonstrate that the method does scale adequately to Industrial Big Data, however further improvements by optimizing the core SVD implementations should still be investigated. 

\subsection{Full diagonal matrices}
\label{subsec:full}

Notation follows \citet[pp. 110--111]{Greenacre_2013}.

\begin{small}
\begin{Verbatim}[commandchars=\\\{\}]
Dr = diag(r);                       
{\color{mygreen}% full diagonal}
Dc = diag(c);                       
{\color{mygreen}% full diagonal}
T = Dr^(-1/2)*(P-r*c')*Dc^(-1/2);   
{\color{mygreen}% standardized residuals (Equation 4)}
[U,S,V] = svd(T,0);                 
{\color{mygreen}% singular value decomposition (economy size)}
F = Dr^(-1/2)*U*S;                  
{\color{mygreen}% principal coordinates of the rows}
\end{Verbatim}
\end{small}

\subsection{Sparse diagonal matrices}
\label{subsec:sparse}

Modifications to overcome the prohibitive size of the diagonal matrices\footnote{Suggested by Odwa Sihlobo, Prescient SA.}.

\begin{small}
\begin{Verbatim}[commandchars=\\\{\}]
rsq = 1./sqrt(r);        
{\color{mygreen}% inverse square root}
nr = 1:numel(r);         
{\color{mygreen}% indices for sparse}
rsp = sparse(nr,nr,rsq); 
{\color{mygreen}% sparse diagonal}
csq = 1./sqrt(c);        
{\color{mygreen}% inverse square root}
nc = 1:numel(c);         
{\color{mygreen}% indices for sparse}
csp = sparse(nc,nc,csq); 
{\color{mygreen}% sparse diagonal}
T = rsp*(P-r*c')*csp;    
{\color{mygreen}% standardized residuals (Equation 4)}
[U,S,V] = svd(T,0);      
{\color{mygreen}% singular value decomposition (economy size)}
F = rsp*U*S;             
{\color{mygreen}% principal coordinates of the rows}
\end{Verbatim}
\end{small}
        
\subsection{Vectorised loops}
\label{subsec:vectprized}

Modifications  to improve speed as well as overcoming the prohibitive size of the diagonal matrices\footnote{Suggested by Stef Salvini, University of Oxford.}.

\begin{small}
\begin{Verbatim}[commandchars=\\\{\}]
rsq = 1./sqrt(r);                   
{\color{mygreen}% inverse square root}
csq = 1./sqrt(c);                   
{\color{mygreen}% inverse square root}
T = P-r*c';                         
for i = 1:size(T,2)                
    T(:,i)=(T(:,i).*rsq(:))*csq(i); 
    {\color{mygreen}% standardized residuals (Equation 4)}
end                                
[U,S,V] = svd(T,0);                 
{\color{mygreen}% singular value decomposition (economy size)}
F = U; s = diag(S);                
for i = 1:size(F,2)                
    F(:,i) = (F(:,i).*rsq(:))*s(i); 
    {\color{mygreen}% principal coordinates of the rows}
end
\end{Verbatim}
\end{small}        
        
\section{\uppercase{Conclusions}}
\label{sec:discussion}

\noindent In this paper we introduce a method for anomaly detection to operate at industrial scale. The method has relevance for system condition monitoring, prognostic health maintenance, or any application where identification of anomalous cases is advantageous. The method aligns with the general trend in Industrial Big Data applications towards data-driven discovery as an alternative to process models which are unlikely to operate at the scales required. The method can accommodate high-dimensional data of heterogeneous data types in a simple and scalable computational framework of well-known and well understood ordination techniques (e.g. SVD, PCA, CA; \cite{Greenacre_2010,Greenacre_2013}), and spatial density analysis (e.g. \cite{Baddeley___2015}).

The method has a well defined workflow (refer to Figure \ref{Fig:Schematic} and to the detailed descriptions in Section \ref{sec:method}) and we demonstrate the method using data from electricity smart meters (refer to Sections \ref{sec:data} and \ref{sec:results} and to Figures \ref{Fig:ConsBiplot} to \ref{Fig:JointRankDensityHistogram}). For the demonstration dataset we find a small number (N\textsubscript{(d$>$2)} = 560; $<$0.5\%) of cases with overwhelming indication that they originate from a mechanism that is different from the one that produced the bulk of cases in the dataset.

This process discovers those cases most resistant to ordination, and therefore least conforming to the mechanism that generated the rest of the data, i.e. the anomalies.

In future work we plan to explore: 

\begin{enumerate}[(i)]

\item the statistical properties of the method, specifically the departure from the mean background process as measured by the joint rank density of cases, 

\item further optimization for operating at scale of the well-known algorithms for singular value decomposition (SVD) as applied in the special case of the method, and 

\item the properties of the method when applying bootstrapped resampling as described in Section~\ref{subsec:random_partitioning}.

\end{enumerate}

\section*{\uppercase{Acknowledgements}}
\label{sec:acknowledgements}
\noindent Funding in support of this work was given by Innovate UK, Grant Reference Number 102510. Useful comments on the manuscript were provided by Laura Shemilt and Glen Moutrie.

\section*{\uppercase{Disclosure}}
\noindent The technical method described in this paper is protected by British patent applications 1713703.5 and 1713896.7, submitted on 25 and 30 August 2017.

\balance

\bibliographystyle{apalike}
{\small
\bibliography{References}}

\vfill
\end{document}